\documentclass{article} %
\usepackage[final]{colm2026_conference}

\usepackage{booktabs}
\usepackage{multirow}
\usepackage{multicol}

\usepackage[x11names]{xcolor}
\usepackage{algorithm}
\usepackage{algorithmic}
\usepackage[most]{tcolorbox}

\usepackage[T1]{fontenc}
\usepackage[utf8]{inputenc}
\usepackage{microtype}

\usepackage{graphicx}
\usepackage{caption}
\usepackage{array}

\usepackage{pgfplots}
\pgfplotsset{compat=1.18}

\renewcommand\cite{\citep}

\usepackage{lineno}

\definecolor{darkblue}{rgb}{0, 0, 0.5}
\usepackage{hyperref}
\usepackage{url}
\hypersetup{colorlinks=true, citecolor=darkblue, linkcolor=darkblue, urlcolor=darkblue}

\title{Smarter by the Moment: Environment-Driven Dynamic \\ Policies for Continual LLM Improvement}

\author{Ting-Wei Chang$^{1}$, Po-Chun Chen$^{1}$, Hen-Hsen Huang$^{2}$ \& Hsin-Hsi Chen$^{1,3}$ \\
  $^{1}$Department of Computer Science and Information Engineering, \\
  \phantom{$^{1}$}National Taiwan University, Taiwan \\
  $^{2}$Institute of Information Science, Academia Sinica, Taiwan \\
  $^{3}$AI Research Center (AINTU), National Taiwan University, Taiwan \\
  \texttt{changtw@nlg.csie.ntu.edu.tw}, \texttt{pcchen@nlg.csie.ntu.edu.tw} \\
  \texttt{hhhuang@iis.sinica.edu.tw}, \texttt{hhchen@ntu.edu.tw}
}

\begin{document}

\ifcolmsubmission
\linenumbers
\fi

\maketitle
\begin{abstract}
Large Language Models (LLMs) have achieved remarkable progress across diverse domains, but continual adaptation to evolving tasks and environments remains a key challenge. Existing memory-augmented approaches retrieve individual past examples as direct references, but do not explicitly synthesize actionable strategies from them, causing the same types of errors to recur. We propose Dynamic Retrieval-based Policy Generation (DRPG), a framework that integrates memory-based retrieval with a dynamic policy generator, leveraging historical data and environment feedback to produce task-specific policies for continual LLM improvement. We evaluate DRPG across six benchmarks spanning text-to-SQL, question answering, medical diagnosis, and Python programming, using seven LLMs from both proprietary and open-weight families. DRPG outperforms strong baselines across most datasets and models. Further analysis demonstrates that DRPG's policy generation is robust to retrieval strategy, operates effectively without prior policy continuity, and can leverage smaller or cross-family models as cost-efficient policy generators. We also find that the benefit of policy-level guidance depends on task characteristics, offering practical insights into when and under what conditions this mechanism is most effective.
\end{abstract}

\section{Introduction}\label{sec:introduction}
Large language models (LLMs) have demonstrated impressive capabilities across a wide range of tasks, from natural language understanding to code generation. However, as real-world environments and requirements evolve, there is a growing need for LLMs to continuously improve their performance and adapt to new challenges~\cite{du2025surveyOntheOptimizationOfLLMbasedAgents, shi2024continual, zheng2025towardslifelonglearning}.\looseness=-1

Among the many adaptation strategies, the online setting is particularly relevant to real-world scenarios, as it requires models to incrementally process new tasks and data streams, reflecting how intelligent systems interact with dynamic environments.
In the online setting, several studies~\cite{hoi2021onlinelearning,rannen2024revisitingonlinefinetuning} update model weights incrementally as new tasks arrive. 
Although this method can be effective, it is not cost-efficient for modern LLMs due to the considerable computational overhead involved. 
As a result, memory-based methods have emerged as promising alternatives. 

For example, GrowPrompt treats the last $k$ processed instances as in-context learning (ICL) examples by directly including them in the prompt, while MemPrompt stores previous questions and answers in an external memory and retrieves the most relevant cases using a retriever~\cite{madaan-etal-2022-memory-MemPrompt-GrowPrompt}. 
Self-StreamICL~\cite{wu2024streambench} further builds on MemPrompt by using only those previous questions that were answered correctly as few-shot references. 
Multi-Agentic-Memory StreamICL (MAM-StreamICL)~\cite{wu2024streambench} takes this concept further by enabling multiple models to share a collective memory, allowing the aggregation of knowledge across agents. 
While these approaches effectively utilize past experiences to guide future predictions, they are limited to providing individual examples as references.

\begin{figure}[t!]
    \centering
    \includegraphics[width=0.95\linewidth]{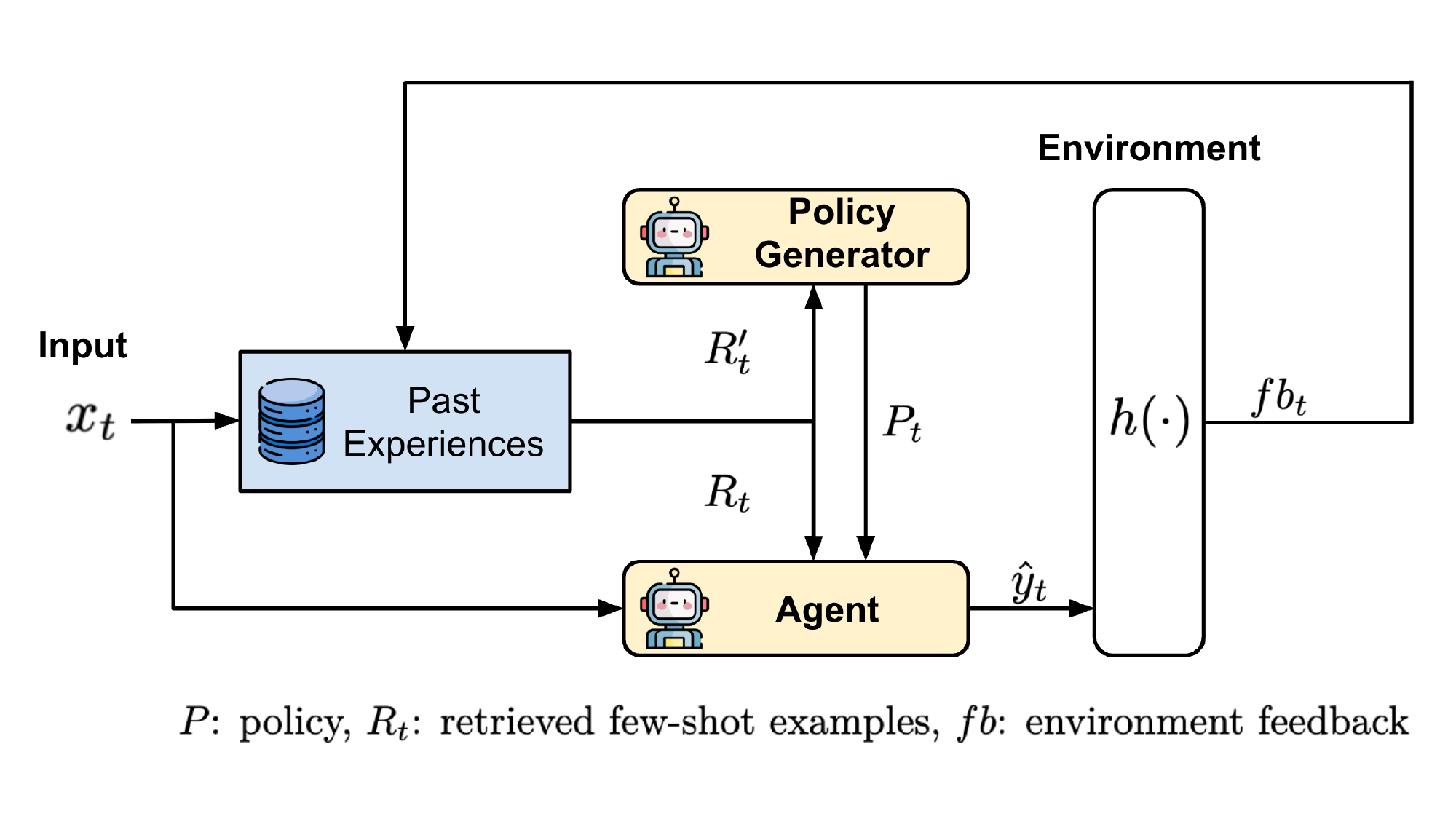}
    \caption{DRPG framework for continual adaptation of LLMs. At each step, the agent predicts using retrieved successful examples and a policy generated by a separate policy generator based on both correct and incorrect prior cases. The environment returns binary feedback, which is stored in memory and used to improve subsequent retrieval and policy generation. DRPG differs from prior methods by leveraging environment feedback to explicitly synthesize policies from past experiences.}
    \label{fig:overview}
\end{figure}

Another direction is the Dynamic Cheatsheet~\cite{suzgun2025dynamiccheatsheet}, which maintains an external memory that is updated during inference, attempting to extract reusable knowledge from previous answers as references for subsequent questions. It has shown effectiveness on several datasets that require multi-step reasoning.
However, it does not leverage environment feedback when deciding what to store, relying solely on the model's own internal signals. 
As a result, its effectiveness is limited when applied to smaller models.\looseness=-1

Despite these efforts, none of these approaches explicitly synthesize feedback-informed strategies that tell the agent what to do differently. As a result, the same types of mistakes may recur even when similar errors have been encountered before.

In this paper, we propose a framework called \textbf{Dynamic Retrieval-based Policy Generation (DRPG)} that addresses this gap by introducing an explicit policy generator that synthesizes actionable strategies, referred to as \emph{policies}, from past experiences and environment feedback. These policies aim to capture recurring patterns across instances, enabling the agent to avoid repeating similar mistakes.

Our main contributions are threefold.
(1) We propose DRPG, a framework that augments memory-based retrieval with a dynamically generated policy for continual LLM adaptation in streaming settings.
(2) Experiments across six benchmarks and seven LLMs show that DRPG outperforms strong baselines across most configurations, with analysis revealing that policy generation is most effective on tasks with systematic, recurring error patterns.
(3) Ablation studies show that DRPG's policy generation is robust to retrieval strategy, operates without prior policy continuity, and allows smaller or cross-family models to serve as policy generators.\looseness=-1

\section{Related Work}
\subsection{LLMs Improving in Online Settings}
Parameter update methods involve updating model weights with each new data point in a streaming setting. \citet{hoi2021onlinelearning} survey online learning algorithms (e.g., Perceptron, Passive--Aggressive), highlighting their iterative nature and theoretical guarantees. 
For LLMs, dynamic evaluation~\citep{rannen2024revisitingonlinefinetuning} adapts model parameters at test time, allowing weights to act as a temporary memory that evolves during inference.

Memory-augmented few-shot methods like GrowPrompt and MemPrompt~\citep{madaan-etal-2022-memory-MemPrompt-GrowPrompt} leverage recent examples and feedback stored in external memory to guide predictions. StreamBench~\citep{wu2024streambench} benchmarks online LLM performance, showing that methods like Self-StreamICL (which use only correct cases as examples) and MAM-StreamICL (which shares memory among models) improve robustness and efficiency.

Dynamic Cheatsheet~\citep{suzgun2025dynamiccheatsheet} introduces a memory buffer for LLMs, dynamically updating stored content based on model confidence and utility, aiming to capture useful intermediate knowledge. Memory management relies solely on internal model signals, not external feedback.

Agent-Pro~\citep{zhang2024agentpro} is an LLM agent that improves via policy-level self-reflection, analyzing action sequences to adjust strategies and outperform static prompt agents in game environments.
SAGE~\citep{liang2024selfSAGE} enhances LLMs through multi-agent iterative feedback, organizing agents in a loop where outputs are refined with self-reflection and memory optimization, leading to continual performance gains.

\subsection{LLMs Improving in Offline or Non-Streaming Settings}
Offline fine-tuning and reinforcement learning methods~\citep{parthasarathy2024ultimateSFT, rafailov2023directDPO, schulman2017proximalPPO} require significant resources and risk catastrophic forgetting~\citep{kalajdzievski2024scalingForget1,song2025completeForget2,li-etal-2024-revisitingForget3, wang2022twoForget4}. LeMa~\citep{an2023learningLEMA} uses GPT-4 for error analysis and correction, transferring mistake correction skills to smaller models through teacher-generated feedback.

Approaches like LEAP~\citep{zhang2024contextLEAP} extract general principles from contrasting correct and incorrect few-shot outputs, which then guide future inference. SALAM~\citep{wang2023learnSALAM} introduces a cooperative ``study assistant'' that analyzes model errors and records them in a ``mistake memory''; during testing, the assistant retrieves similar errors to provide targeted guidelines that help the agent anticipate and avoid recurring mistakes. In a related vein, Induct-Learn~\citep{chen-etal-2024-induct} induces task-level pseudo instructions from a small number of correct demonstrations and a short task phrase, then combines these instructions with demonstrations to guide the LLM's problem-solving process at inference time. Similarly, \citet{chen-etal-2025-diverge} propose generating diverse reasoning rationales and then inducing a unified strategy, demonstrating the effectiveness of induction-based approaches for LLM reasoning. These methods use data-driven reflections to generate reusable guidance. However, they operate in offline or static settings and do not continuously adapt from environment feedback in a streaming manner.

Reflexion~\citep{shinn2023reflexion} enables agents to learn from feedback stored in episodic and long-term memory, but it requires repeated interactions with the same dataset. Self-Refine~\citep{madaan2023selfRefine} uses iterative self-feedback to correct mistakes. While this method enhances output quality across diverse tasks, it does not use ground truth as external feedback, which can lead to over-refinement of correct responses or prematurely halting refinement on incorrect ones.

In contrast to these lines of work, the novelty of DRPG lies less in its individual components than in the capability their combination enables: continually synthesizing reusable, cross-query policies from environment feedback within a stream. Reflection methods such as Reflexion repeatedly critique a single instance and do not accumulate strategies across queries, while offline induction methods such as LEAP and Induct-Learn produce fixed guidance once; neither continually distills feedback-grounded, transferable strategies as the stream evolves.

\section{Dynamic Retrieval-based Policy Generation (DRPG) Framework}

In this section, we introduce the DRPG framework (Figure~\ref{fig:overview}). At each time step, the framework operates as follows: (1) the agent retrieves relevant correct examples from memory; (2) a separate policy generator retrieves both correct and incorrect examples and synthesizes them into a concise set of actionable guidelines (the \emph{policy}); (3) the agent produces an answer conditioned on both the retrieved examples and the policy; (4) the environment provides binary feedback, which is stored in memory for future use. Below we formalize each component.

\textbf{Agent.} The agent is modeled as a function $f(\cdot)$. At each time step $t$, given an input question $x_t$, the agent extracts a set of top-$k$ few-shot examples
\begin{equation}
    R_t = r\left(x_t, \left\{ (x_i, \hat{y}_i, fb_i) \mid i < t,\, fb_i = 1 \right\},\, k \right)
\end{equation}
where $fb_i \in \{0, 1\}$ is the binary correctness feedback provided by the environment for the $i$-th past example, $R_t$ denotes the set of top-$k$ most relevant previous examples with positive feedback ($fb_i = 1$), retrieved from the agent's past interactions before time $t$, and $r(\cdot)$ is the retriever. The agent then produces an output as
\begin{equation}
    \hat{y}_t = f(x_t, R_t, P_t)
\end{equation}
where $P_t$ is the policy at time $t$.

\textbf{Policy Generator.} The policy generator $g(\cdot)$ is responsible for producing the policy $P$, a concise set of actionable guidelines (e.g., up to five bullet points) distilled from past correct and incorrect examples. At each time step $t$, its inputs include the retrieved examples $R'_t$ and it produces the policy as
\begin{equation}
    P_t = g\left(R'_t\right)
    \label{eq:policy_generator}
\end{equation}

We use a retriever $r'(\cdot)$ to retrieve top-$k$ relevant examples from previous experiences based on the current input $x_t$. The retriever $r'(\cdot)$ may be the same as or different from the agent's retriever $r(\cdot)$. In this work, we adopt a contrastive setting as the default configuration, where the retrieval is performed as follows:
\begin{equation}
    R'_{t,1} = r'\left(x_t, \left\{(x_i, \hat{y}_i, fb_i) \mid i < t,\, fb_i = 1 \right\}, k/2\right)
\end{equation}
\begin{equation}
    R'_{t,0} = r'\left(x_t, \left\{(x_j, \hat{y}_j, fb_j) \mid j < t,\, fb_j = 0 \right\}, k/2\right)
\end{equation}
We then form the final retrieved set by combining these two subsets:
\begin{equation}
    R'_t = R'_{t,1} \cup R'_{t,0}
\end{equation}

\textbf{Environment interaction.} The environment, denoted as $h(\cdot)$, produces the binary correctness feedback introduced above: after the agent outputs $\hat{y}_t$, the environment returns $fb_t = h(x_t, \hat{y}_t) \in \{0, 1\}$, indicating the correctness of the answer.

\textbf{Past Experiences.} During the streaming process, we maintain an updatable memory database of past experiences. At time step $t$, this database contains tuples of the form $(x_i, \hat{y}_i, fb_i)$ for $i = 1, \ldots, t-1$.

The full procedure is summarized in Algorithm~\ref{alg:algorithm-drpg}.

\begin{algorithm}[h]
\small
\caption{Framework for DRPG}
\label{alg:algorithm-drpg}
\begin{algorithmic}[1]
\STATE Initialize agent $f(\cdot)$, retrievers $r(\cdot)$, $r'(\cdot)$, and policy generator $g(\cdot)$
\FOR{$t = 1$ to $T$}
    \STATE Receive instance $x_t$ from the data stream;
    \STATE Retrieve\\
    $R_t = r\left(x_t, \left\{(x_i, \hat{y}_i, fb_i) \mid i < t,\, fb_i = 1 \right\}, k\right)$;
    \STATE \textbf{// Contrastive retrieval:}
    \STATE $R'_{t,1} = r'\left(x_t, \left\{(x_i, \hat{y}_i, fb_i) \mid i < t,\, fb_i = 1 \right\}, \frac{k}{2} \right)$;
    \STATE $R'_{t,0} = r'\left(x_t, \left\{(x_j, \hat{y}_j, fb_j) \mid j < t,\, fb_j = 0 \right\}, \frac{k}{2} \right)$;
    \STATE $R'_t = R'_{t,1} \cup R'_{t,0}$; 
    \STATE The policy generator generates policy $P_t = g\left(R'_t\right)$;
    \STATE The agent predicts $\hat{y}_t = f(x_t, R_t, P_t)$;
    \STATE Receive feedback $fb_t = h(x_t, \hat{y}_t)$, $fb_t \in \{0, 1\}$;
    \STATE Store triplet $(x_t, \hat{y}_t, fb_t)$ in memory;
\ENDFOR
\end{algorithmic}
\end{algorithm}

\subsection{Prompt Design}
We follow the prompt design of StreamBench~\cite{wu2024streambench}. To minimize the impact of prompt engineering, we use the same prompt structure across different methods. This consistent structure is applied to both the agent and the policy generator.

The main components of our prompt structure are as follows, with certain parts added or omitted depending on the specific method:

(1)~\textbf{Role assignment}, which specifies the role of the LLM for the given task;

(2)~\textbf{Reference materials}, including task-related supporting information such as answer choices, few-shot examples from memory, or policies;

(3)~\textbf{Question}, the current input ($x_t$) the agent is required to answer; and

(4)~\textbf{Additional instructions}, covering formatting requirements, rules, or restrictions specific to the task.

For the policy generator, we use a prompt template that includes role assignment, reference materials (retrieved correct and incorrect examples), and additional instructions. 
The policy generator analyzes past cases and produces a policy accordingly, focusing on actionable bullet points relevant to the task. 
The detailed prompts for each dataset can be found in Appendix~\ref{app:prompts}.
We adapt the prompt to dataset-specific content while retaining the same prompt structure.

\section{Experiments}
\subsection{Datasets}
Following StreamBench~\cite{wu2024streambench}, we adopt most of the datasets used in their benchmark, excluding ToolBench since it uses an LLM as a judge, which heavily relies on the LLM's own judgment ability.
Our evaluation spans four task categories.

\textbf{Text-to-SQL.} Spider~\cite{yu2018spider}, CoSQL~\cite{yu2019cosql}, and BIRD~\cite{li2023canbird} are all large-scale, cross-domain datasets with complex SQL queries, designed to test generalization to unseen database schemas. These three benchmarks vary in difficulty: Spider is the easiest, while BIRD is the most challenging.

\textbf{Question Answering.} HotpotQA~\cite{yang2018hotpotqa} is a multi-hop question answering (QA) dataset that requires the LLM to locate and reason over supporting passages. The dataset uses a distractor setting, in which the provided passages contain both useful and irrelevant information.

\textbf{Medical Diagnosis.} DDXPlus is a synthetic dataset that includes patient profiles and full differential diagnoses, simulating scenarios in which a doctor diagnoses a patient's condition~\citep{fansi2022ddxplus}. The agent needs to select the most appropriate diagnosis from 49 candidate options, based on the patient's background.

\textbf{Python Programming.} DS-1000~\cite{lai2023ds} provides 1,000 real-world Python programming tasks across seven libraries (e.g., NumPy, pandas), with perturbations to avoid memorization and support reliable execution-based evaluation.

\subsection{Evaluation Metrics}

We adopt the standard evaluation metric for each dataset. For the Text-to-SQL datasets, we use the common execution accuracy metric, which compares the execution results of the generated SQL queries with the ground truth. For HotpotQA, we follow the original paper and use exact match as the primary metric, comparing the agent's answer to the ground truth. DDXPlus is treated as a 49-choice multiple-choice task, so we use accuracy as the evaluation metric. For Python programming tasks, we use the standard pass@1 metric, in which the agent generates only one answer and the result is judged by running several predefined test cases.

\subsection{Baselines}

We compare DRPG against the following approaches, including both non-streaming and streaming methods.

\textbf{Zero-shot} evaluates the base ability of each LLM without any additional adaptation.

\textbf{Self-Refine}~\citep{madaan2023selfRefine} allows the agent to generate an initial answer and then refine it using self-generated feedback, considering only the current question and answer without external memory.

\textbf{Self-StreamICL}~\citep{wu2024streambench} stores each question, answer, and environment feedback in an external memory $\mathcal{M}$, then retrieves similar and correctly answered past cases as few-shot examples for new questions. It outperforms both GrowPrompt and MemPrompt, and serves as the main baseline in our study.

For DRPG, we use the same agent configuration as Self-StreamICL, but additionally include a dynamically generated policy in the agent's prompt. The policy generator retrieves $k$ relevant past examples for each update. In our default configuration, inspired by prior work~\cite{gao2024customizingconstrastive, an2023learningLEMA, zhang2024contextLEAP}, we adopt a contrastive setting that retrieves an equal number of correct and incorrect cases; however, as shown in our ablation study (Appendix~\ref{app:retrieval-strategy}), policy generation is robust across different retrieval strategies.

\subsection{Models}
To assess the generalizability of DRPG, we evaluate across three model families spanning both proprietary API-based and open-weight LLMs: Gemini (\texttt{gemini-2.0-flash}, \texttt{gemini-2.0-flash-lite})~\cite{Anil2023GeminiAF}, Llama (\texttt{llama-3.3-70b}~\cite{dubey2024llama}, \texttt{llama-4-maverick}, \texttt{llama-4-scout}~\cite{meta2025llama4blog}), and Mistral (\texttt{mistral-medium}~\cite{mistral_medium3}, \texttt{mistral-small}~\cite{mistral_small3.1_2503}). This selection covers a range of model sizes and capability levels, enabling us to examine whether the benefits of policy generation are consistent across different architectures and scales. In addition, Appendix~\ref{app:qwen} reports results on two additional open-weight families, Qwen and Gemma. While this spread covers multiple families and scales, it does not by itself establish generality across all modern LLM ecosystems (e.g., OpenAI or Claude models); extending the evaluation to further families is left to future work.

\subsection{Experimental Setup}

In this study, we fix several parameters to ensure experimental consistency, following many settings from StreamBench. First, the retriever uses the BAAI/bge-base-en-v1.5~\cite{bge_embedding} embedding model. For the number of few-shot examples, we follow the agent settings in StreamBench~\cite{wu2024streambench}: the retriever selects $k$ past instances as ICL few-shot examples. Specifically, $k=16$ is used for Spider, CoSQL, BIRD, and DDXPlus, while $k=4$ is used for DS-1000 and HotpotQA to avoid exceeding the context window. The policy generator uses the same $k$ value as the agent for retrieval. For dataset ordering, we use a fixed random seed (42) to ensure consistent shuffling across all experiments. Additionally, we set the temperature to 0 to maintain consistency and the maximum output tokens to 1,024 for all experiments. We fully reuse StreamBench's retrieval pipeline without modification, shared by DRPG and all baselines; its precise configuration is given in Appendix~\ref{app:retrieval-details}.

\section{Results and Analysis}
This section presents a comprehensive analysis of the DRPG framework. We first establish its overall performance against strong baselines (Section~\ref{sec:main-performance}), then examine the relationship between task characteristics and the effectiveness of policy-level guidance (Section~\ref{sec:case-study}). Finally, we conduct a series of ablation studies examining the robustness of the policy generation mechanism (Section~\ref{sec:ablation}).

\subsection{Performance of DRPG Framework}
\label{sec:main-performance}

\begin{table}[t!]
\centering
\small

\setlength{\tabcolsep}{2pt}

\begin{tabular}{l l rrr r r r}
\toprule
& & \multicolumn{3}{c}{\textbf{Text-to-SQL}} & \multicolumn{1}{c}{\textbf{QA}} & \multicolumn{1}{c}{\textbf{Medical}} & \multicolumn{1}{c}{\textbf{Python}} \\
\cmidrule(lr){3-5} \cmidrule(lr){6-6} \cmidrule(lr){7-7} \cmidrule(lr){8-8}
\textbf{Model} & \textbf{Method} & \textbf{Spider} & \textbf{CoSQL} & \textbf{BIRD} & \textbf{HotpotQA} & \textbf{DDXPlus} & \textbf{DS-1000} \\
\midrule
\multirow{4}{*}{\textbf{gemini-2.0-flash}} 
& Zero-shot    & 81.00 & 58.29 & 43.42 & 61.20 & 68.99 & \textbf{80.70} \\
& Self-Refine      & 81.14 & 58.29 & 43.81 & 60.80 & 68.65 & 80.50 \\
& Self-StreamICL   & 83.47 & \textbf{63.56} & \textbf{46.61} & 64.07 & 89.51 & 79.20 \\
& DRPG (Ours)      & \textbf{83.65} & 63.06 & 45.63 & \textbf{65.07} & \textbf{91.38} & 79.70 \\
\midrule
\multirow{4}{*}{\textbf{gemini-2.0-flash-lite}} 
& Zero-shot    & 80.76 & 57.80 & 40.61 & 50.93 & 60.37 & 74.40 \\
& Self-Refine      & 80.67 & \textbf{58.09} & 40.74 & 50.80 & 60.03 & 74.20 \\
& Self-StreamICL   & 77.92 & 43.89 & \textbf{42.18} & \textbf{58.47} & \textbf{90.25} & 74.70 \\
& DRPG (Ours)      & \textbf{83.65} & 52.93 & 41.92 & 57.53 & 88.10 & \textbf{75.70} \\
\midrule
\multirow{4}{*}{\textbf{llama-3.3-70b}} 
& Zero-shot    & 47.97 & 44.29 & 9.78 & 63.53 & 48.47 & 72.40 \\
& Self-Refine      & 76.90 & 48.16 & 32.33 & 63.33 & 61.90 & 71.40 \\
& Self-StreamICL   & 76.06 & 60.77 & 33.77 & 63.53 & \textbf{74.32} & \textbf{73.60} \\
& DRPG (Ours)      & \textbf{81.18} & \textbf{61.87} & \textbf{41.98} & \textbf{63.87} & 70.24 & 72.90 \\
\midrule
\multirow{4}{*}{\textbf{llama-4-maverick}} 
& Zero-shot    & 43.92 & 41.81 & 8.15 & 63.80 & 16.78 & 71.90 \\
& Self-Refine      & 70.80 & 54.62 & 31.68 & \textbf{64.47} & 29.88 & \textbf{74.00} \\
& Self-StreamICL   & 74.80 & 59.78 & 36.05 & 62.73 & 72.11 & 71.90 \\
& DRPG (Ours)      & \textbf{79.69} & \textbf{60.87} & \textbf{42.18} & 50.60 & \textbf{84.07} & 70.90 \\
\midrule
\multirow{4}{*}{\textbf{llama-4-scout}} 
& Zero-shot    & 38.38 & 38.63 & 6.91 & 57.33 & 37.36 & \textbf{32.60} \\
& Self-Refine      & 69.49 & 52.04 & 31.68 & \textbf{58.47} & 59.30 & 22.40 \\
& Self-StreamICL   & 68.33 & 55.91 & 31.23 & 55.87 & \textbf{61.05} & 18.40 \\
& DRPG (Ours)      & \textbf{76.85} & \textbf{57.20} & \textbf{39.50} & 56.67 & 50.79 & 23.30 \\
\midrule
\multirow{4}{*}{\textbf{mistral-medium}} 
& Zero-shot    & 39.60 & 39.82 & 8.21 & 65.00 & 55.05 & 71.80 \\
& Self-Refine      & 69.90 & 52.93 & 29.14 & 65.07 & 67.63 & 72.70 \\
& Self-StreamICL   & 73.60 & 57.89 & 35.59 & 65.93 & \textbf{88.04} & \textbf{74.10} \\
& DRPG (Ours)      & \textbf{78.20} & \textbf{58.79} & \textbf{38.79} & \textbf{66.93} & 86.51 & 70.80 \\
\midrule
\multirow{4}{*}{\textbf{mistral-small-2503}} 
& Zero-shot    & 40.29 & 39.92 & 7.69 & 62.13 & 59.92 & 67.90 \\
& Self-Refine      & 67.91 & 52.53 & 27.64 & 62.73 & 58.22 & 66.20 \\
& Self-StreamICL   & 72.57 & 57.60 & 31.03 & 62.33 & 78.63 & \textbf{72.10} \\
& DRPG (Ours)      & \textbf{78.16} & \textbf{58.59} & \textbf{33.51} & \textbf{64.93} & \textbf{81.24} & 68.90 \\
\midrule[\heavyrulewidth]
\multirow{3}{*}{\textbf{\shortstack{DRPG\\Win Count}}} 
& vs Zero-shot & 7/7 & 6/7 & 7/7 & 5/7 & 7/7 & 3/7 \\
& vs Self-Refine & 7/7 & 6/7 & 7/7 & 5/7 & 6/7 & 4/7 \\
& vs Self-StreamICL & 7/7 & 6/7 & 5/7 & 5/7 & 3/7 & 3/7 \\
\bottomrule
\end{tabular}
\caption{Performance comparison of DRPG (Ours) and baseline methods across multiple LLMs and benchmark tasks. The best-performing method for each model-task pair is highlighted in \textbf{bold}. The bottom rows show pairwise win counts of DRPG against each baseline across all models.}
\label{tab:main-results}
\end{table}

Table~\ref{tab:main-results} presents the main comparison between DRPG and existing baselines across six benchmarks and seven LLMs. DRPG outperforms Self-StreamICL with a pairwise win record of 29 out of 42 model--dataset combinations. The margins are even more pronounced against Self-Refine (35/42) and Zero-shot (35/42), highlighting the overall advantage of DRPG across a wide range of tasks and model families.

The most consistent gains are observed on text-to-SQL benchmarks (Spider, CoSQL, and BIRD), where DRPG wins against Self-StreamICL in 18 out of 21 model configurations. DRPG also outperforms Self-StreamICL in 5 out of 7 configurations on the multi-hop reasoning benchmark HotpotQA. A notable exception is \texttt{llama-4-maverick}, which shows a significant drop on HotpotQA; we hypothesize that the generated policy occasionally overrides the agent's correct reasoning on multi-hop questions.

On DDXPlus and DS-1000, DRPG wins against Self-StreamICL in only 3 out of 7 configurations for each dataset. We observe that on these tasks, the generated policies tend to capture overly narrow, instance-specific patterns rather than broadly applicable strategies, limiting their coverage. Treating each model--dataset configuration as a pair, DRPG significantly outperforms Self-StreamICL overall (one-sided Wilcoxon signed-rank test, $p = 0.005$), while on DDXPlus and DS-1000 the differences are not significant in either direction (two-sided $p = 0.94$ and $0.73$; Appendix~\ref{app:significance} reports all tests and visualizes the per-configuration gains). DRPG's benefit is therefore not uniform across task types: it concentrates on tasks whose errors share recurring, generalizable structure, while on tasks requiring instance-specific knowledge DRPG performs on par with, rather than above, Self-StreamICL. We examine this task-dependent behavior in detail below.

\subsection{Task-Dependent Effectiveness of Policy Generation}
\label{sec:case-study}
We qualitatively examine the policies generated by DRPG to understand \emph{why} policy generation is more effective on some tasks than others. The key distinction lies in whether a task exhibits error patterns that can be captured by a small number of high-level rules generalizing across instances.

\textbf{Text-to-SQL and QA: policies capture generalizable rules.} On Spider, the generated policy for \texttt{llama-4-scout} (Figure~\ref{fig:policy_spider_scout}) captures actionable, cross-instance guidelines such as ``Validate Schema and Relationships'' and ``Avoid Ambiguous or Redundant Results.'' Each rule applies broadly across different SQL queries, enabling an 8.5 percentage point improvement over Self-StreamICL. This is possible because text-to-SQL errors are structurally regular: mistakes like incorrect joins, missing aggregation, or schema misuse recur regardless of the specific query. A similar pattern holds for HotpotQA (5/7 wins), where multi-hop reasoning involves recurring challenges such as identifying the correct supporting passages and avoiding distractor information.

\begin{figure}[t!]
    \centering
    \fbox{
  \begin{minipage}{0.97\linewidth}
  \small
- Directly Address the Question: Focus on directly answering the question asked, avoiding unnecessary joins or conditions that do not contribute to answering the query.

- Validate Schema and Relationships: Verify the schema and relationships between tables to ensure correct joins, subqueries, and conditions are used.

- Precise Use of SQL Constructs: Choose SQL constructs that accurately handle the query requirements, such as using aggregation or grouping when necessary, and handle edge cases like empty result sets or division by zero.

- Avoid Ambiguous or Redundant Results: Ensure that queries are clear and unambiguous, directly answering the question without providing redundant or unnecessary information.

- Handle Multiple Results Correctly: Ensure correct usage of subqueries, aggregations, and grouping to handle complex queries with multiple results, considering all required columns and potential duplicates.
  \end{minipage}
}
    \caption{Example of the generated policy at time step 100 on \texttt{llama-4-scout} for the Spider dataset. Each rule is broadly applicable across different SQL queries.}
\label{fig:policy_spider_scout}
\end{figure}

\textbf{Medical Diagnosis and Python Programming: policies become too narrow.} In contrast, the policies generated for DDXPlus (Figure~\ref{fig:policy_ddxplus_scout}) tend to focus on specific diseases, such as ``Consider Cardiovascular Diseases for Chest Pain'' or ``Evaluate for Anaphylaxis in Allergic Reactions.'' With 49 candidate diagnoses and diverse symptom profiles, the retrieved examples often span unrelated disease categories, making it difficult to extract a coherent diagnostic strategy. Unlike text-to-SQL, where structural patterns (e.g., correct use of JOINs) generalize across queries, medical diagnosis lacks a universal procedure that applies across diverse conditions; the policy generator therefore defaults to enumerating disease-specific heuristics, each covering only a narrow subset of future cases. Similarly, DS-1000 spans seven Python libraries with distinct API conventions; the precise, library-specific knowledge required (e.g., the correct parameters for \texttt{pandas.pivot\_table} or \texttt{numpy.reshape}) cannot be effectively compressed into five actionable bullet points. For such tasks, instance-level retrieval, which provides concrete, directly relevant examples, remains the more effective mechanism.

\begin{figure}[t!]
    \centering
    \fbox{%
        \begin{minipage}{0.97\linewidth}
        \small
        - Consider Cardiovascular Diseases for Chest Pain: When patients report chest pain, especially if it's described as tedious, heavy, or sharp, and radiates to areas like the biceps, shoulders, or under the jaw, consider cardiovascular diseases such as NSTEMI/STEMI, especially in patients with risk factors like diabetes, high cholesterol, smoking, or family history of cardiovascular diseases.

        - Evaluate for Anaphylaxis in Allergic Reactions: In cases of known severe food allergies, recent consumption of allergenic substances, symptoms like swelling, redness, itching, and widespread skin lesions, consider anaphylaxis, especially if accompanied by respiratory distress or cardiovascular symptoms.

        - Assess for Infectious Diseases Based on Exposure and Symptoms: For patients with recent travel history to high-risk areas (e.g., West Africa for Ebola), contact with infected individuals, symptoms like fever, shortness of breath, and diffuse muscle pain, consider infectious diseases such as Ebola.
        \end{minipage}
    }
    \caption{Partial example of the generated policy at time step 100 on \texttt{llama-4-scout} for DDXPlus. Unlike Spider, rules focus on specific diseases rather than generalizable strategies.}
    \label{fig:policy_ddxplus_scout}
\end{figure}

These observations suggest that policy generation is most beneficial for tasks with \emph{compositional structure} and \emph{recurring error types}, and less so for tasks relying on instance-specific knowledge or broad categorical recall.

\subsection{Ablation Studies}
\label{sec:ablation}

We examine the sensitivity of DRPG to key design choices, focusing on the model used for the policy generator.

In DRPG, the agent and the policy generator can be instantiated using different language models. Table~\ref{tab:teacher} presents the results under various model pairings, including both within-series and cross-series configurations.

Overall, the agent's own capability remains the dominant factor determining final performance, while the policy generator provides auxiliary guidance that leads to performance improvements in most cases. Notably, even when using a smaller LLM as the policy generator, the induced strategies still benefit a stronger agent. An interesting case is \texttt{maverick}: when it generates its own policy, performance drops sharply on HotpotQA (50.60), but when \texttt{scout} or \texttt{mistral-small} serves as the policy generator, this drop does not occur (63.73 and 63.07, respectively). This suggests that cross-model policy generation can sometimes avoid failure modes present in same-model generation, though we note that for \texttt{scout} and \texttt{mistral-small} as agents, same-model and cross-model configurations perform comparably.

These results suggest that lightweight models can serve as cost-efficient policy generators without sacrificing downstream performance, and that practitioners may benefit from experimenting with cross-model configurations.

\begin{table}[t!]
\centering
\small

\setlength{\tabcolsep}{1.5pt}

\begin{tabular}{l l l rrr r r r}
\toprule
& &  \multicolumn{1}{l}{\textbf{Policy}}& \multicolumn{3}{c}{\textbf{Text-to-SQL}} & \multicolumn{1}{c}{\textbf{QA}} & \multicolumn{1}{c}{\textbf{Medical}} & \multicolumn{1}{c}{\textbf{Python}} \\
\cmidrule(lr){4-6} \cmidrule(lr){7-7} \cmidrule(lr){8-8} \cmidrule(lr){9-9}
\textbf{Method} & \textbf{Agent} & \multicolumn{1}{l}{\textbf{Generator}} & \textbf{Spider} & \textbf{CoSQL} & \textbf{BIRD} & \textbf{HotpotQA} & \textbf{DDXPlus} & \textbf{DS-1000} \\
\midrule
\multirow{3}{*}{\textbf{Self-StreamICL}} 
& maverick & -- & 74.80 & 59.78 & 36.05 & 62.73 & 72.11 & 71.90 \\
& scout & -- & 68.33 & 55.91 & 31.23 & 55.87 & 61.05 & 18.40 \\
& mistral-small & -- & 72.57 & 57.60 & 31.03 & 62.33 & 78.63 & 72.10 \\
\midrule
\multirow{7}{*}{\textbf{DRPG (Ours)}} 
& maverick & maverick & 79.69 & 60.87 & \textbf{42.18} & 50.60 & 84.07 & 70.90 \\
& maverick & scout & \textbf{81.70} & \textbf{61.07} & 42.05 & 63.73 & \textbf{84.86} & \textbf{72.90} \\
& maverick & mistral-small & 79.93 & 60.38 & 41.66 & 63.07 & 83.96 & 70.00 \\
\cmidrule(lr){2-9}
& scout & maverick & 76.67 & 55.51 & 37.55 & 56.53 & 60.60 & 31.50 \\
& scout & scout & 76.85 & 57.20 & 39.50 & 56.67 & 50.79 & 23.30 \\
\cmidrule(lr){2-9}
& mistral-small & maverick & 77.69 & 58.49 & 35.01 & 64.60 & 82.60 & 69.30 \\
& mistral-small & mistral-small & 78.16 & 58.59 & 33.51 & \textbf{64.93} & 81.24 & 68.90 \\
\bottomrule
\end{tabular}
\caption{Performance when using different models for the main agent and the policy generator. For Self-StreamICL, there is no separate policy generator. We abbreviate Llama-4-Maverick as \textit{maverick}, Llama-4-Scout as \textit{scout}, and Mistral-Small-2503 as \textit{mistral-small}. The best-performing combination is highlighted in \textbf{bold}.}

\label{tab:teacher}
\end{table}

We further verify two additional design choices. First, replacing the contrastive retrieval strategy with correct-only or wrong-only retrieval yields comparable performance, confirming that the policy generator can reliably distill strategies regardless of the composition of its input (Appendix~\ref{app:retrieval-strategy}). Second, providing the previous policy $P_{t-1}$ as input to the policy generator does not consistently improve results, indicating that the mechanism operates effectively in a stateless manner without requiring policy continuity (Appendix~\ref{app:previous-policy}). Together with the cross-model result above, these findings demonstrate that DRPG's policy generation is a modular component with minimal sensitivity to design choices. In addition, a no-feedback ablation in Appendix~\ref{app:no-feedback} isolates the contribution of environment feedback: generating policies from the retrieved examples without correctness labels consistently underperforms DRPG across all six benchmarks.

\subsection{Cost Analysis}
\label{sec:cost-analysis}

Relative to Self-StreamICL, DRPG adds one retrieval (fetching the successful and failed cases for policy generation) and one LLM call (the policy generator) per query. Table~\ref{tab:cost-main} compares the per-query computational cost of the three methods for the representative configuration \texttt{mistral-medium} $\times$ Spider (N=2,147); token counts include all of a method's LLM calls. Because most runs were executed on free-tier APIs, the latency should be read only as a relative comparison; its ordering is consistent with the call structure: DRPG ($\approx$2 calls) $>$ Self-StreamICL $>$ Zero-shot. Per-query token counts for all 42 (model $\times$ dataset) configurations are provided in Appendix~\ref{app:cost}.

\begin{table}[h]
\centering
\small
\setlength{\tabcolsep}{5pt}
\begin{tabular}{l cccc}
\toprule
\textbf{Method} & \textbf{Retrievals/query} & \textbf{LLM calls/query} & \textbf{Avg tokens (in/out)} & \textbf{Avg latency} \\
\midrule
Zero-shot      & 0 & 1 & 162 / 13    & $\approx$ 1.8 s \\
Self-StreamICL & 1 & 1 & 585 / 17    & $\approx$ 5.3 s \\
DRPG (Ours)    & 2 & 2 & 1,676 / 143 & $\approx$ 9.3 s \\
\bottomrule
\end{tabular}
\caption{Per-query cost comparison for the representative configuration \texttt{mistral-medium} $\times$ Spider (N=2,147). Token counts include all LLM calls of each method; latency is measured on free-tier APIs and should be read as a relative comparison.}
\label{tab:cost-main}
\end{table}

\section{Conclusion}
We present DRPG, a framework that augments memory-based retrieval with a dynamic policy generator for continual LLM improvement in online settings. Evaluation across six benchmarks and seven LLMs shows that DRPG outperforms strong baselines on the majority of model--dataset configurations, significantly so overall against Self-StreamICL, with the largest gains on structured prediction tasks, while performing on par with Self-StreamICL on tasks that require instance-specific knowledge. Ablation studies further demonstrate that policy generation is robust to retrieval strategy, operates statelessly, and allows the use of a separate, potentially smaller model as the policy generator. Our analysis also reveals that the benefit of policy-level guidance varies with task characteristics: it is most effective for tasks with systematic, recurring error patterns capturable by high-level rules, and less so for tasks requiring instance-specific knowledge. We believe this finding offers practical guidance for practitioners and future work on when to deploy policy-level adaptation mechanisms in streaming settings.

\section*{Acknowledgments}
This work was supported by the National Science and Technology Council, Taiwan, under grants NSTC 114-2221-E-002-070-MY3 and NSTC 115-2634-F-002-012, as well as by financial support from the Featured Area Research Center Program within the framework of the Higher Education Sprout Project by the Ministry of Education (115L900901).

\bibliography{custom}
\bibliographystyle{colm2026_conference}
\clearpage
\appendix

\section*{Limitations}
Our evaluation adopts a controlled online setting with clean binary correctness feedback at every step, identical to the StreamBench protocol used by Self-StreamICL and the other baselines, so that all methods are compared under the same conditions and the observed behavior can be attributed to policy generation itself. We note that per-step binary correctness feedback is an idealization: even in text-to-SQL, execution can reveal syntactic errors but does not by itself confirm that the produced query is semantically correct, and fully reliable correctness signals generally require additional verification such as user confirmation or downstream checks. In real deployments we expect such feedback to be sparse rather than available at every step; a natural extension of DRPG is to update memory and generate policies only from the subset of interactions that do receive feedback, and how its advantage over feedback-free methods depends on feedback density is an open question. Our conclusions hold for this controlled setting and should not be extrapolated to real-deployment conditions such as non-stationary streams or noisy, delayed, partial, non-binary, or sparse feedback; studying robustness under such degraded-feedback environments is left to future work. In addition, while we use a consistent prompt structure across all methods, model families, and datasets, we have not conducted a systematic sensitivity study over prompt paraphrases, and we leave establishing prompt robustness to future work.

Due to budget limitations, we did not evaluate DRPG with reasoning LLMs such as OpenAI GPT-o3 or DeepSeek R1, which we leave to future work. Although our benchmarks span four diverse task categories, the effectiveness of DRPG may vary on other tasks. All experiments use a single fixed data ordering (seed=42) with temperature set to 0; sensitivity to data order in streaming settings remains unexplored, and we do not report variance across multiple runs. We also observe that the generated policy can occasionally hurt performance, as seen with \texttt{llama-4-maverick} on HotpotQA; understanding when policy-level guidance may conflict with the agent's reasoning is an important direction for future investigation. We do not include Dynamic Cheatsheet~\cite{suzgun2025dynamiccheatsheet} as a baseline, since its implementation supports only specific datasets and does not cover the benchmarks we use; as a closer controlled alternative, Appendix~\ref{app:no-feedback} compares DRPG with a no-environment-feedback variant of policy generation.

\label{sec:appendix}
\section{Use of Large Language Models}
ChatGPT was used for grammar refinement and occasional assistance in programming tasks. All language model outputs served only as references; the final text and code were written entirely by the authors.

\section{Experimental Cost}
\label{app:cost}

All experiments conducted in this study were performed within the free-tier allocations provided by the respective API providers. Specifically, we utilized the complimentary access quotas offered by Google AI Studio, Nvidia NIM (NVIDIA Inference Microservices), and Meta Llama API. These free-tier limits were sufficient for our experimental requirements. Consequently, no additional computational expenses or API usage fees were incurred during the course of this research.

Table~\ref{tab:cost-perquery} complements the representative comparison in Section~\ref{sec:cost-analysis} by reporting the per-query input/output token counts for all 42 (model $\times$ dataset) configurations.

\begin{figure}[t]
    \centering
    \includegraphics[width=0.95\linewidth]{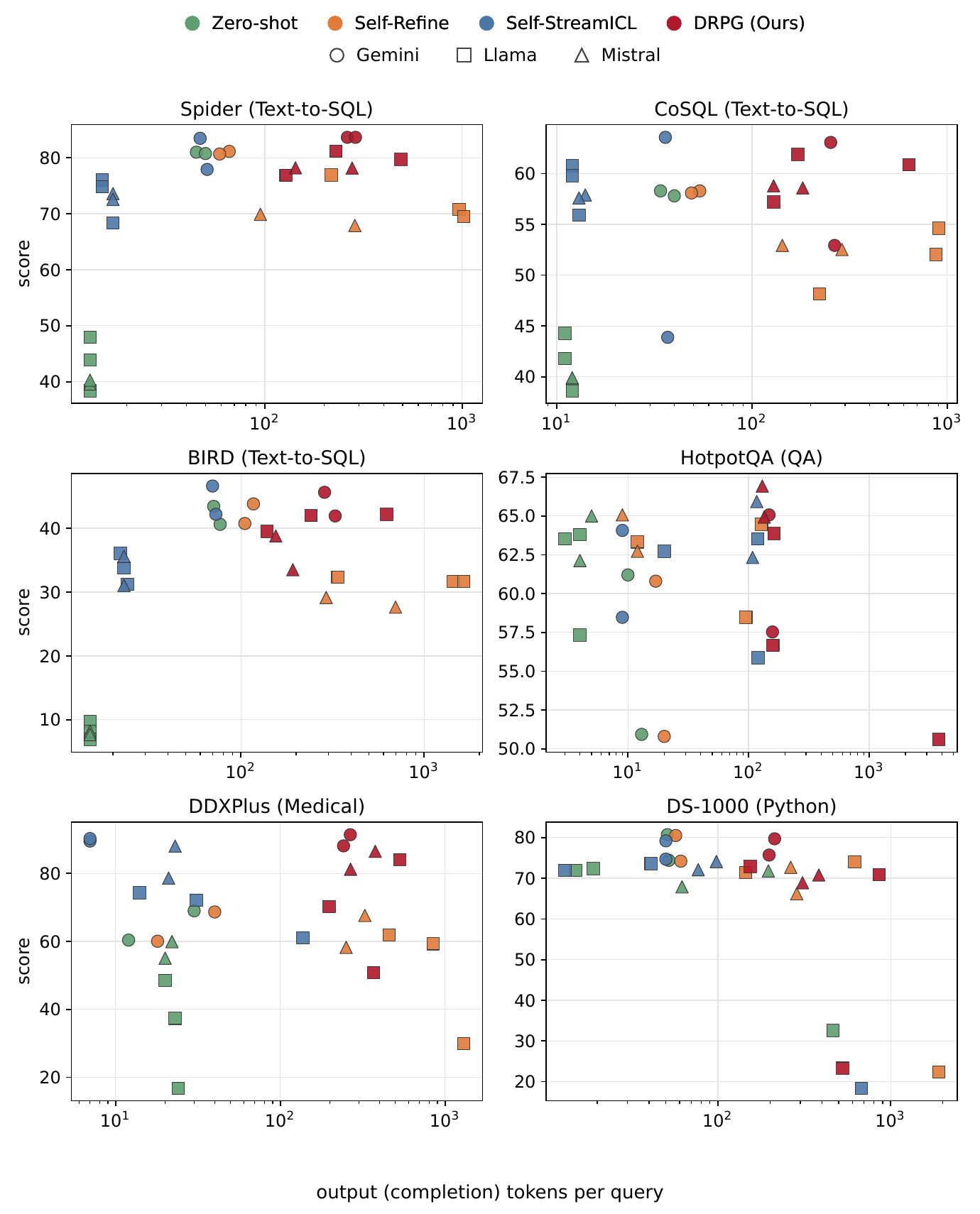}
    \caption{Score versus per-query output (completion) tokens for all 42 model--dataset configurations and four methods (scores from Table~\ref{tab:main-results}, token counts from Table~\ref{tab:cost-perquery}). Each point is one model under one method; token counts include all LLM calls of the method. Up and left is better.}
    \label{fig:acc-vs-token}
\end{figure}

Figure~\ref{fig:acc-vs-token} further visualizes this cost--performance trade-off, plotting the score of each configuration against its per-query \emph{output} (completion) tokens. We focus on output tokens because autoregressive decoding makes them the dominant factor in wall-clock latency, whereas input tokens are consumed in a single parallel prefill pass~\citep{pope2023efficiently, patel2024splitwise}. Relative to Self-Refine, DRPG generally sits toward the upper left: higher scores at comparable or lower output cost. Relative to Self-StreamICL, DRPG spends additional output tokens (mainly for policy generation), and whether this cost translates into gains depends on the task type, consistent with the analysis in Section 5: clear improvements on text-to-SQL, comparable performance on tasks relying on instance-specific knowledge.

\begin{table}[t]
\centering
\setlength{\tabcolsep}{3pt}
\resizebox{\textwidth}{!}{%
\begin{tabular}{ll cccccc}
\toprule
 & & \multicolumn{3}{c}{\textbf{Text-to-SQL}} & \textbf{QA} & \textbf{Medical} & \textbf{Python} \\
\cmidrule(lr){3-5}
\textbf{Model} & \textbf{Method} & \textbf{Spider} & \textbf{CoSQL} & \textbf{BIRD} & \textbf{HotpotQA} & \textbf{DDXPlus} & \textbf{DS-1000} \\
\midrule
\multirow{4}{*}{\textbf{gemini-2.0-flash}} & Zero-shot & 409/45 & 450/34 & 928/71 & 1,488/10 & 645/30 & 1,273/51 \\
 & Self-Refine & 960/66 & 1,025/54 & 2,153/117 & 3,125/17 & 1,391/40 & 2,636/57 \\
 & Self-StreamICL & 1,293/47 & 1,263/36 & 2,199/70 & 7,201/9 & 4,582/7 & 2,776/50 \\
 & DRPG (Ours) & 2,947/262 & 2,794/253 & 4,302/286 & 13,275/148 & 9,322/266 & 4,717/213 \\
\midrule
\multirow{4}{*}{\textbf{gemini-2.0-flash-lite}} & Zero-shot & 409/50 & 450/40 & 928/77 & 1,488/13 & 645/12 & 1,273/52 \\
 & Self-Refine & 928/59 & 1,011/49 & 2,116/105 & 3,028/20 & 1,352/18 & 3,495/61 \\
 & Self-StreamICL & 1,408/51 & 1,345/37 & 2,215/73 & 7,186/9 & 4,590/7 & 2,744/50 \\
 & DRPG (Ours) & 3,238/288 & 2,856/265 & 4,494/327 & 13,283/158 & 9,289/242 & 4,692/198 \\
\midrule
\multirow{4}{*}{\textbf{llama-3.3-70b}} & Zero-shot & 162/13 & 170/11 & 392/15 & 1,017/3 & 346/20 & 385/19 \\
 & Self-Refine & 443/217 & 1,012/222 & 1,617/338 & 2,503/12 & 1,555/457 & 947/145 \\
 & Self-StreamICL & 574/15 & 572/12 & 932/23 & 6,967/119 & 3,075/14 & 2,324/41 \\
 & DRPG (Ours) & 2,642/229 & 1,553/172 & 3,771/241 & 12,922/163 & 6,345/198 & 2,005/154 \\
\midrule
\multirow{4}{*}{\textbf{llama-4-maverick}} & Zero-shot & 162/13 & 170/11 & 392/15 & 1,017/4 & 346/24 & 402/15 \\
 & Self-Refine & 2,594/966 & 2,535/904 & 6,561/1,446 & 2,996/128 & 4,192/1,300 & 2,501/620 \\
 & Self-StreamICL & 575/15 & 557/12 & 925/22 & 6,918/20 & 2,991/31 & 1,053/13 \\
 & DRPG (Ours) & 1,684/489 & 2,451/637 & 3,804/625 & 16,613/3,768 & 6,292/530 & 4,234/858 \\
\midrule
\multirow{4}{*}{\textbf{llama-4-scout}} & Zero-shot & 162/13 & 170/12 & 392/15 & 1,017/4 & 346/23 & 385/464 \\
 & Self-Refine & 2,631/1,020 & 2,302/876 & 6,234/1,647 & 2,128/95 & 2,232/845 & 7,353/1,897 \\
 & Self-StreamICL & 571/17 & 552/13 & 935/24 & 6,886/120 & 3,085/137 & 2,657/677 \\
 & DRPG (Ours) & 1,690/128 & 1,523/129 & 2,185/139 & 12,713/159 & 6,384/368 & 2,373/527 \\
\midrule
\multirow{4}{*}{\textbf{mistral-medium}} & Zero-shot & 162/13 & 170/12 & 392/15 & 1,017/5 & 346/20 & 385/196 \\
 & Self-Refine & 428/95 & 998/143 & 2,414/292 & 2,061/9 & 1,593/326 & 1,014/264 \\
 & Self-StreamICL & 585/17 & 555/14 & 913/23 & 7,309/117 & 3,047/23 & 1,021/98 \\
 & DRPG (Ours) & 1,676/143 & 1,505/129 & 2,148/155 & 9,034/130 & 8,889/377 & 4,941/384 \\
\midrule
\multirow{4}{*}{\textbf{mistral-small-2503}} & Zero-shot & 162/13 & 170/12 & 392/15 & 1,017/4 & 346/22 & 385/62 \\
 & Self-Refine & 1,557/286 & 1,619/289 & 4,552/698 & 2,376/12 & 1,898/251 & 1,588/286 \\
 & Self-StreamICL & 582/17 & 553/13 & 901/23 & 7,315/108 & 2,983/21 & 2,640/77 \\
 & DRPG (Ours) & 2,828/277 & 1,540/182 & 2,143/192 & 9,026/135 & 8,945/267 & 4,646/309 \\
\bottomrule
\end{tabular}}%
\caption{Per-query computational cost (input/output tokens per query, i.e.\ total tokens divided by the number of questions). Each DRPG value already includes both of its LLM calls (policy generation and answer); each Self-Refine value includes all refinement rounds.}
\label{tab:cost-perquery}
\end{table}

\section{Retrieval Pipeline Details}
\label{app:retrieval-details}

We fully reuse StreamBench's retrieval pipeline without any modification; DRPG and Self-StreamICL share this exact, unmodified retriever, so retrieval is not a component tuned for DRPG. Table~\ref{tab:retrieval-details} summarizes the precise configuration.

\begin{table}[h]
\centering
\small
\begin{tabular}{@{}l >{\raggedright\arraybackslash}p{8.2cm}@{}}
\toprule
\textbf{Item} & \textbf{Setting} \\
\midrule
Embedding model & BAAI/bge-base-en-v1.5 (CLS pooling, L2-normalized) \\
\addlinespace
Similarity metric & Cosine similarity, top-$k$ (implemented as FAISS nearest-neighbour search over the L2-normalized embeddings, whose ranking is equivalent to cosine similarity) \\
\addlinespace
Embedded key & The input question only \\
\addlinespace
Task-specific inputs & SQL schema, QA context, and code test cases are provided via the prompt template and are not included in the embedded key \\
\addlinespace
Setup across datasets & Identical; only $k$ differs \\
\addlinespace
$k$ & 16 (Spider, CoSQL, BIRD, DDXPlus); 4 (HotpotQA, DS-1000), to fit the context window \\
\addlinespace
Policy generator & Uses the same $k$, balanced evenly between correct and incorrect cases \\
\bottomrule
\end{tabular}
\caption{Precise configuration of the retrieval pipeline, fully reused from StreamBench and shared by DRPG and all baselines.}
\label{tab:retrieval-details}
\end{table}

\section{Robustness to Retrieval Strategy}
\label{app:retrieval-strategy}

To examine the sensitivity of DRPG to the retrieval strategy used for policy generation, we compare three configurations: contrastive (our default), which retrieves an equal number of correct and incorrect examples; correct-only, which retrieves only successful cases; and wrong-only, which uses only failed cases.

As shown in Table~\ref{tab:retrieval-method}, all three strategies achieve comparable performance across benchmarks and models. This indicates that the policy generation mechanism can reliably distill useful strategies from past experiences, regardless of whether the input examples are successes, failures, or a mixture of both. The policy generator is not sensitive to the composition of its input, suggesting that the act of synthesizing high-level rules is itself the key driver of improvement, rather than the specific examples provided. This also challenges the common practice of discarding failure cases in memory-based methods~\cite{wu2024streambench, min2022rethinkingnegative1, wei2023largernegative2}, consistent with recent findings that analyzing errors can yield generalizable insights~\cite{zhang2024contextLEAP}.

We note the scope of this ablation: it varies the \emph{composition} of the retrieved cases (contrastive, correct-only, wrong-only), not the relevance computation itself. We did not ablate the embedding model, similarity metric, embedded fields, or $k$, since the retriever (Appendix~\ref{app:retrieval-details}) is shared and fixed across DRPG and all baselines and is not a component we optimized; a systematic relevance-computation ablation is left to future work.

\begin{table}[h]
\centering
\small
\setlength{\tabcolsep}{2pt}

\begin{tabular}{l l rrr r r r}
\toprule
& & \multicolumn{3}{c}{\textbf{Text-to-SQL}} & \multicolumn{1}{c}{\textbf{QA}} & \multicolumn{1}{c}{\textbf{Medical}} & \multicolumn{1}{c}{\textbf{Python}} \\
\cmidrule(lr){3-5} \cmidrule(lr){6-6} \cmidrule(lr){7-7} \cmidrule(lr){8-8}
\textbf{Model} & \textbf{Retrieval Strategy} & \textbf{Spider} & \textbf{CoSQL} & \textbf{BIRD} & \textbf{HotpotQA} & \textbf{DDXPlus} & \textbf{DS-1000} \\
\midrule
\multirow{3}{*}{\textbf{gemini-2.0-flash}} 
& Contrastive    & \textbf{83.65} & \textbf{63.06} & 45.63 & \textbf{65.07} & \textbf{91.38} & 79.70 \\
& Wrong cases    & 83.51 & 62.66 & \textbf{47.26} & 64.93 & 88.44 & 80.40 \\
& Correct cases  & 83.42 & 61.97 & 45.96 & 64.13 & 88.04 & \textbf{80.80} \\
\midrule
\multirow{3}{*}{\textbf{gemini-2.0-flash-lite}} 
& Contrastive    & \textbf{83.65} & 52.93 & 41.92 & 57.53 & 88.10 & \textbf{75.70} \\
& Wrong cases    & 83.09 & \textbf{55.71} & \textbf{43.81} & 57.40 & 83.50 & 74.40 \\
& Correct cases  & 82.49 & 49.45 & 43.61 & \textbf{57.67} & \textbf{90.42} & 74.70 \\
\midrule
\multirow{3}{*}{\textbf{llama-4-maverick}} 
& Contrastive    & 79.69 & \textbf{60.87} & \textbf{42.18} & 50.60 & 84.07 & 70.90 \\
& Wrong cases    & 79.51 & 57.30 & 41.98 & 60.87 & 80.22 & \textbf{72.10} \\
& Correct cases  & \textbf{79.79} & 56.70 & 40.74 & \textbf{63.00} & \textbf{84.47} & 69.60 \\
\midrule
\multirow{3}{*}{\textbf{llama-4-scout}} 
& Contrastive    & \textbf{76.85} & 57.20 & \textbf{39.50} & \textbf{56.67} & 50.79 & 23.30 \\
& Wrong cases    & 76.25 & \textbf{57.70} & 37.29 & \textbf{56.67} & 54.42 & 24.70 \\
& Correct cases  & 74.94 & 57.40 & 38.14 & 56.40 & \textbf{58.45} & \textbf{25.80} \\
\midrule
\multirow{3}{*}{\textbf{mistral-small-2503}} 
& Contrastive    & 78.16 & 58.59 & 33.51 & 64.93 & \textbf{81.24} & 68.90 \\
& Wrong cases    & 77.78 & 57.80 & \textbf{36.70} & \textbf{65.00} & 78.74 & \textbf{69.40} \\
& Correct cases  & \textbf{78.20} & \textbf{58.69} & 36.18 & 64.60 & \textbf{81.24} & 67.40 \\
\bottomrule
\end{tabular}
\caption{Performance of different retrieval strategies in policy generation. The ``Contrastive'' method uses an equal number of past correct and incorrect few-shot examples. The best-performing method for each model-task pair is highlighted in \textbf{bold}.}
\label{tab:retrieval-method}
\end{table}

\section{Effect of Referencing the Previous Policy}
\label{app:previous-policy}

In our default DRPG design, the policy generator constructs each policy independently at every time step. To examine whether maintaining policy continuity across steps is beneficial, we introduce a variant in which the policy generator receives $P_{t-1}$ as additional input when generating $P_t$.

Table~\ref{tab:previous-policy} reports the full results. Neither configuration consistently outperforms the other across models and tasks, indicating that the policy generator can effectively synthesize useful strategies from retrieved examples alone, without requiring access to prior policies. This stateless property simplifies system design and avoids the risk of error accumulation from inheriting outdated or overly specific rules, while the agent's growing memory of past experiences (Eq.~1) still ensures continual adaptation at the framework level.

\begin{table}[h]
\centering
\small
\setlength{\tabcolsep}{2pt}

\begin{tabular}{l c rrr r r r}
\toprule
& & \multicolumn{3}{c}{\textbf{Text-to-SQL}} & \multicolumn{1}{c}{\textbf{QA}} & \multicolumn{1}{c}{\textbf{Medical}} & \multicolumn{1}{c}{\textbf{Python}} \\
\cmidrule(lr){3-5} \cmidrule(lr){6-6} \cmidrule(lr){7-7} \cmidrule(lr){8-8}
\textbf{Model} & \textbf{Previous Policy} & \textbf{Spider} & \textbf{CoSQL} & \textbf{BIRD} & \textbf{HotpotQA} & \textbf{DDXPlus} & \textbf{DS-1000} \\
\midrule
\multirow{2}{*}{\textbf{gemini-2.0-flash}} 
& w/    & \textbf{83.79} & \textbf{63.46} & \textbf{46.68} & 62.33 & 90.93 & \textbf{81.60} \\
& w/o   & 83.65 & 63.06 & 45.63 & \textbf{65.07} & \textbf{91.38} & 79.70 \\
\midrule
\multirow{2}{*}{\textbf{gemini-2.0-flash-lite}} 
& w/    & 82.35 & \textbf{53.23} & \textbf{43.48} & 55.60 & 86.22 & 75.20 \\
& w/o   & \textbf{83.65} & 52.93 & 41.92 & \textbf{57.53} & \textbf{88.10} & \textbf{75.70} \\
\midrule
\multirow{2}{*}{\textbf{llama-4-maverick}} 
& w/    & \textbf{81.04} & 60.58 & 41.66 & \textbf{63.33} & 82.03 & \textbf{71.20} \\
& w/o   & 79.69 & \textbf{60.87} & \textbf{42.18} & 50.60 & \textbf{84.07} & 70.90 \\
\midrule
\multirow{2}{*}{\textbf{llama-4-scout}} 
& w/    & 76.53 & 56.60 & 37.68 & 56.53 & \textbf{59.92} & \textbf{25.70} \\
& w/o   & \textbf{76.85} & \textbf{57.20} & \textbf{39.50} & \textbf{56.67} & 50.79 & 23.30 \\
\midrule
\multirow{2}{*}{\textbf{mistral-small-2503}} 
& w/    & 77.64 & 57.20 & \textbf{35.33} & \textbf{65.93} & 78.51 & \textbf{72.10} \\
& w/o   & \textbf{78.16} & \textbf{58.59} & 33.51 & 64.93 & \textbf{81.24} & 68.90 \\
\bottomrule
\end{tabular}
\caption{Ablation study on the impact of using the previous turn's policy. ``w/'' denotes with previous policy, ``w/o'' denotes without. The best-performing setting for each model-task pair is highlighted in \textbf{bold}.}
\label{tab:previous-policy}
\end{table}

\section{Effect of Environment Feedback: A No-Feedback Ablation}
\label{app:no-feedback}

To isolate the contribution of environment feedback, we construct a no-feedback variant of DRPG, in which the policy is generated entirely from the retrieved few-shot examples without any correct/incorrect labels; following Dynamic Cheatsheet, this variant uses no environment feedback. All other components (the policy generator, prompts, retriever, $k$, and seed) are identical to the DRPG configuration in Table~\ref{tab:main-results}. Due to cost constraints, we run this ablation with \texttt{llama-3.3-70b} on all six benchmarks.

As shown in Table~\ref{tab:no-feedback}, DRPG outperforms its no-feedback variant on all six datasets, indicating that the gains come specifically from feedback-conditioned policy generation rather than from adding policy text or summarizing retrieved examples. The gap is largest on DDXPlus and DS-1000, where policies generated without feedback fall far below Self-StreamICL; consistent with the task-dependent analysis in Section~5, DRPG does not surpass Self-StreamICL on these two datasets even with feedback, while feedback keeps its performance close to that baseline.

\begin{table}[h]
\centering
\setlength{\tabcolsep}{4pt}
\begin{tabular}{l rrr r r r}
\toprule
& \multicolumn{3}{c}{\textbf{Text-to-SQL}} & \multicolumn{1}{c}{\textbf{QA}} & \multicolumn{1}{c}{\textbf{Medical}} & \multicolumn{1}{c}{\textbf{Python}} \\
\cmidrule(lr){2-4} \cmidrule(lr){5-5} \cmidrule(lr){6-6} \cmidrule(lr){7-7}
\textbf{Method} & \textbf{Spider} & \textbf{CoSQL} & \textbf{BIRD} & \textbf{HotpotQA} & \textbf{DDXPlus} & \textbf{DS-1000} \\
\midrule
Self-StreamICL     & 76.06 & 60.77 & 33.77 & 63.53 & \textbf{74.32} & \textbf{73.60} \\
DRPG w/o feedback  & 78.81 & 60.08 & 35.07 & 63.73 & 57.43 & 61.50 \\
DRPG (Ours)        & \textbf{81.18} & \textbf{61.87} & \textbf{41.98} & \textbf{63.87} & 70.24 & 72.90 \\
\bottomrule
\end{tabular}
\caption{No-feedback ablation with \texttt{llama-3.3-70b} (same configuration as Table~\ref{tab:main-results}). ``DRPG w/o feedback'' generates the policy from the retrieved examples without correctness labels. The best result per dataset is highlighted in \textbf{bold}.}
\label{tab:no-feedback}
\end{table}

\section{Significance Tests and Per-Configuration Gains}
\label{app:significance}

Treating each model--dataset configuration in Table~\ref{tab:main-results} as a paired observation (DRPG vs.\ Self-StreamICL; $n = 7$ models per dataset, 42 configurations overall), we run paired Wilcoxon signed-rank tests. Across all 42 configurations, DRPG significantly outperforms Self-StreamICL (one-sided $p = 0.005$). Table~\ref{tab:significance} reports the per-dataset two-sided tests: the improvements are significant on Spider and CoSQL, while on DDXPlus and DS-1000 the differences are not significant in either direction, with mean differences of only $-0.2$ and $-0.3$ points. In other words, DRPG does not significantly degrade performance on any task type while improving significantly overall.

\begin{table}[h]
\centering
\small
\begin{tabular}{l r r}
\toprule
\textbf{Dataset} & \textbf{Two-sided $p$} & \textbf{Mean $\Delta$ (DRPG $-$ Self-StreamICL)} \\
\midrule
Spider   & 0.016 & $+4.9$ \\
CoSQL    & 0.031 & $+2.0$ \\
BIRD     & 0.078 & $+3.9$ \\
HotpotQA & 0.547 & $-1.0$ \\
DDXPlus  & 0.938 & $-0.2$ \\
DS-1000  & 0.734 & $-0.3$ \\
\bottomrule
\end{tabular}
\caption{Paired Wilcoxon signed-rank tests of DRPG vs.\ Self-StreamICL per dataset ($n = 7$ models each), computed from Table~\ref{tab:main-results}. Overall, across all 42 configurations, DRPG is significantly better (one-sided $p = 0.005$).}
\label{tab:significance}
\end{table}

\begin{figure}[h]
    \centering
    \includegraphics[width=0.95\linewidth]{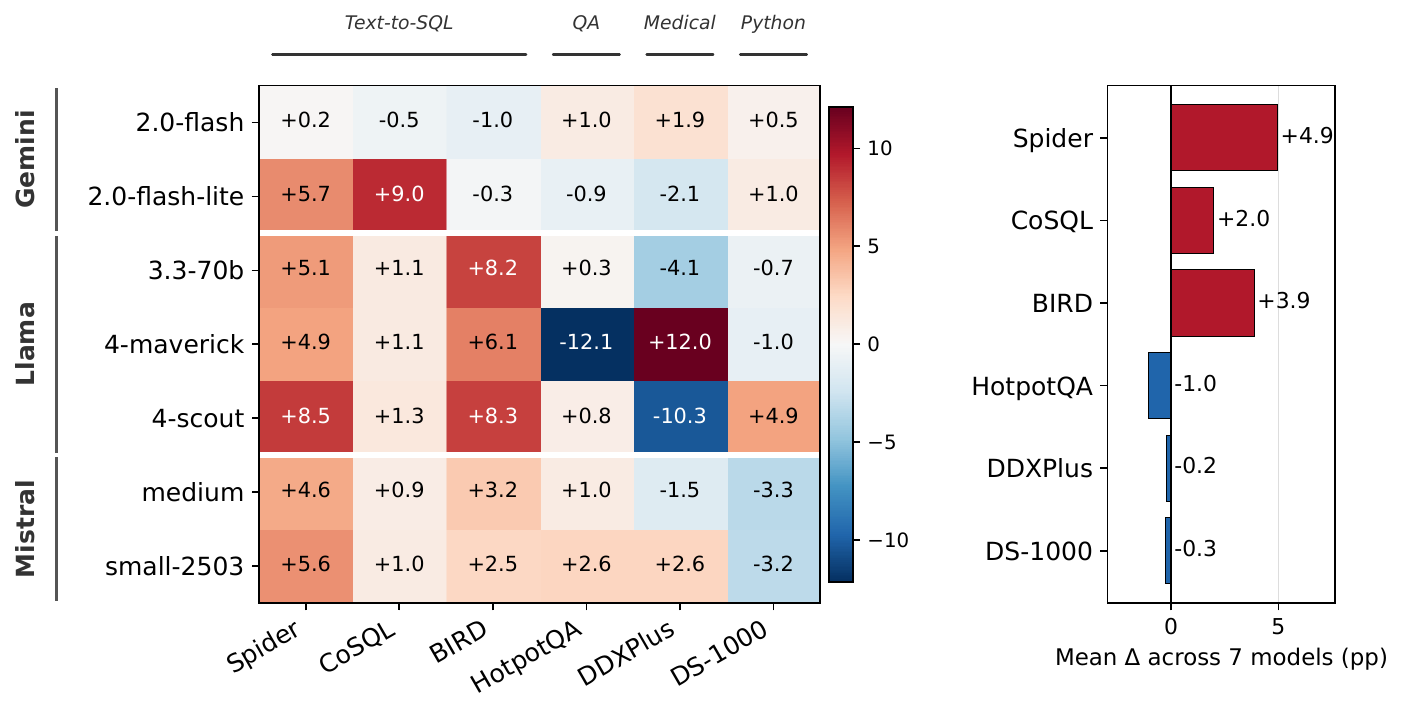}
    \caption{$\Delta$(DRPG $-$ Self-StreamICL), in percentage points, for every model--dataset configuration, grouped by model family (computed from Table~\ref{tab:main-results}; shown to one decimal).}
    \label{fig:delta-heatmap}
\end{figure}

Figure~\ref{fig:delta-heatmap} visualizes the per-configuration gains $\Delta$(DRPG $-$ Self-StreamICL), grouped by model family. The gains concentrate on the text-to-SQL benchmarks and HotpotQA, consistent with the task-dependent analysis in Section~5.

\section{Results on Additional Open-Weight Model Families: Qwen and Gemma}
\label{app:qwen}

To further examine generality beyond the three model families in Table~\ref{tab:main-results}, we evaluate \texttt{qwen3.5-122b-a10b} (Qwen)~\cite{qwenteam2026qwen35} and \texttt{gemma-4-31b-it} (Gemma)~\cite{gemmateam2026gemma4} on all six benchmarks, comparing Zero-shot, the strongest baseline Self-StreamICL, and DRPG under the same experimental setup as Table~\ref{tab:main-results}; the results are shown in Table~\ref{tab:qwen}. Both models are hybrid reasoning models; we disable their thinking mode in all runs (thinking budget set to 0, so no reasoning traces are generated), consistent with our scope of evaluating standard, non-reasoning inference as stated in the Limitations.

For Qwen, DRPG outperforms Self-StreamICL on Spider, BIRD, HotpotQA, and DS-1000, performs comparably on CoSQL, and falls below Self-StreamICL on DDXPlus. This is consistent with the task-dependent pattern analyzed in Section~5: policy-level guidance helps most on tasks whose errors share recurring, generalizable patterns, whereas on DDXPlus the generated policies tend to capture overly narrow, instance-specific patterns rather than broadly applicable strategies, limiting their benefit.

For Gemma, DRPG outperforms Self-StreamICL on five of the six datasets and is comparable on HotpotQA. Notably, Self-StreamICL falls well below Zero-shot on BIRD and DS-1000 for this model, while DRPG does not exhibit the same degradation: it stays above Zero-shot on BIRD and recovers most of the gap on DS-1000. This suggests that policy-level guidance can be more robust than accumulating raw exemplars for some models.

\begin{table}[h]
\centering
\small
\setlength{\tabcolsep}{3pt}
\begin{tabular}{l l rrr r r r}
\toprule
& & \multicolumn{3}{c}{\textbf{Text-to-SQL}} & \multicolumn{1}{c}{\textbf{QA}} & \multicolumn{1}{c}{\textbf{Medical}} & \multicolumn{1}{c}{\textbf{Python}} \\
\cmidrule(lr){3-5} \cmidrule(lr){6-6} \cmidrule(lr){7-7} \cmidrule(lr){8-8}
\textbf{Model} & \textbf{Method} & \textbf{Spider} & \textbf{CoSQL} & \textbf{BIRD} & \textbf{HotpotQA} & \textbf{DDXPlus} & \textbf{DS-1000} \\
\midrule
\multirow{3}{*}{\textbf{qwen3.5-122b-a10b}}
& Zero-shot      & 69.59 & 50.43 & 32.50 & 54.75 & 54.37 & 57.48 \\
& Self-StreamICL & 69.91 & \textbf{51.97} & 32.10 & 55.16 & \textbf{75.74} & 58.58 \\
& DRPG (Ours)    & \textbf{70.33} & 51.34 & \textbf{35.72} & \textbf{57.04} & 65.80 & \textbf{62.24} \\
\midrule
\multirow{3}{*}{\textbf{gemma-4-31b-it}}
& Zero-shot      & 75.08 & 56.60 & 36.70 & 63.27 & 68.20 & \textbf{78.10} \\
& Self-StreamICL & 75.50 & 57.50 & 27.51 & \textbf{65.27} & 90.02 & 64.60 \\
& DRPG (Ours)    & \textbf{78.48} & \textbf{57.89} & \textbf{38.01} & 64.80 & \textbf{90.76} & 72.80 \\
\bottomrule
\end{tabular}
\caption{Performance of \texttt{qwen3.5-122b-a10b} and \texttt{gemma-4-31b-it} across the six benchmarks, evaluated under the same protocol as Table~\ref{tab:main-results} with the thinking mode of both models disabled (thinking budget set to 0). The best result per model--dataset pair is highlighted in \textbf{bold}.}
\label{tab:qwen}
\end{table}

\section{Prompt Templates for the Agent and Policy Generator}
\label{app:prompts}

This section presents the prompt templates used in our DRPG framework. We provide agent prompts (Figures~\ref{fig:prompt_text2sql}--\ref{fig:prompt_ds1000}) that guide task execution with few-shot examples and policies, and policy generator prompts (Figures~\ref{fig:policy_text2sql}--\ref{fig:policy_ds1000}) that refine policies from error analysis. The templates cover text-to-SQL, question answering, medical diagnosis, and code generation tasks.

\begin{figure}[t!]
  \centering
  \small
  \renewcommand{\arraystretch}{1.1}
  \begin{tabular}{p{0.93\textwidth}}
    \toprule
    \textbf{[Role assignment]} \\
    You are performing the text-to-SQL task. \\
    \\
    \textbf{[Reference materials]} \\
    Here are some examples: \\
    \textcolor{blue}{\{few-shot examples\}} \\
    \\
    Please pay special attention to the following points, which are derived from previous cases. \\
    \textcolor{blue}{\{policy\}} \\
    \\
    \textbf{[Question]} \\
    Now it's your turn. \\
    - SQL schema: \textcolor{blue}{\{schema\}} \\
    - Using valid SQLite, answer the following question for the SQL schema provided above. \\
    - Question: \textcolor{blue}{\{question\}} \\
    \\
    \textbf{[Additional instructions]} \\
    Now, generate the correct SQL code directly (Do NOT generate other text except the SQL code): \\
    \texttt{'''sql\textbackslash n<your SQL code>\textbackslash n'''} \\
    \bottomrule
  \end{tabular}
  \vspace{-0.7em}
  \caption{The prompt template for the DRPG for text-to-SQL. The \textcolor{blue}{\{schema\}} would be replaced by database schema based on dataset.}
  \label{fig:prompt_text2sql}
\end{figure}

\begin{figure}[t!]
  \centering
  \small
  \renewcommand{\arraystretch}{1.1}
  \begin{tabular}{p{0.93\textwidth}}
    \toprule
    \textbf{[Role assignment]} \\
    You are doing a question-answering task. \\
    \\
    \textbf{[Reference materials]} \\
    Here are some example cases: \textcolor{blue}{\{few-shot examples\}} \\
    \\
    Please pay special attention to the following points, which are derived from previous cases. \\
    \textcolor{blue}{\{policy\}} \\
    \\
    \textbf{[Question]} \\
    Now you are given the following context, which might help you answer the question: \\
    Context: \textcolor{blue}{\{context\}} \\
    \\
    Question: \textcolor{blue}{\{question\}} \\
    \\
    \textbf{[Additional instructions]} \\
    Note that you only need to answer with a short text span without explanation. Now, provide your answer in the following JSON format: \\
    \texttt{\{"answer": "<your answer text span>"\}} \\
    \bottomrule
  \end{tabular}
  \vspace{-0.7em}
  \caption{The prompt template for the DRPG for HotpotQA. The \textcolor{blue}{\{context\}} would be replaced by provided passages.}
  \label{fig:prompt_hotpotqa}
\end{figure}

\begin{figure}[t!]
  \centering
  \small
  \renewcommand{\arraystretch}{1.1}
  \begin{tabular}{p{0.93\textwidth}}
    \toprule
    \textbf{[Role assignment]} \\
    Act as a medical doctor and diagnose the patient based on the provided patient profile. \\
    \\
    \textbf{[Reference materials]} \\
    All possible diagnoses for you to choose from are as follows (one diagnosis per line, in the format of <number>. <diagnosis>): \\
    \textcolor{blue}{\{option text\}} \\
    \\
    Here are some example cases. \textcolor{blue}{\{few-shot examples\}} \\
    \\
    Please pay special attention to the following points, which are derived from previous cases. \\
    \textcolor{blue}{\{policy\}} \\
    \\
    \textbf{[Question]} \\
    Now it's your turn. \\
    \textcolor{blue}{\{profile\}} \\
    \\
    \textbf{[Additional instructions]} \\
    Now, directly provide the diagnosis for the patient in the following format: \\
    \texttt{<number>. <diagnosis>} \\
    \bottomrule
  \end{tabular}
  \vspace{-0.7em}
  \caption{The prompt template for the DRPG used in DDXPlus. The \textcolor{blue}{\{profile\}} is replaced with the provided patient profile, and the \textcolor{blue}{\{option text\}} contains all possible diagnoses.}
  \label{fig:prompt_ddxplus}
\end{figure}

\begin{figure}[t!]
  \centering
  \small
  \renewcommand{\arraystretch}{1.1}
  \begin{tabular}{p{0.93\textwidth}}
    \toprule
    \textbf{[Role assignment]} \\
    You are performing a python programming task to satisfy the user's requirements. \\
    \\
    \textbf{[Reference materials]} \\
    Here are some examples: \\
    \textcolor{blue}{\{few-shot examples\}} \\
    \\
    Please pay special attention to the following points, which are derived from previous cases. \\
    \textcolor{blue}{\{policy\}} \\
    \\
    \textbf{[Question]} \\
    Now it's your turn. \\
    The user's requirements (enclosed in "''): \\
    \textcolor{blue}{\{question\}} \\
    \\
    \textbf{[Additional instructions]} \\
    You need to provide your solution in python code to satisfy the user's requirements. Your code will be tested as follows (enclosed in "''): \\
    \textcolor{blue}{\{test cases\}} \\
    \\
    Now, generate your code directly in the following format: \\
    \texttt{'''python} \\
    \texttt{<your code>} \\
    \texttt{'''} \\
    \bottomrule
  \end{tabular}
  \vspace{-0.7em}
  \caption{The prompt template for the DRPG used in DS-1000. The \textcolor{blue}{\{question\}} is replaced with the user's requirements, and \textcolor{blue}{\{test cases\}} is the provided execution context for code testing.}
  \label{fig:prompt_ds1000}
\end{figure}

\begin{figure}[t!]
  \centering
  \small
  \renewcommand{\arraystretch}{1.1}
  \begin{tabular}{p{0.93\textwidth}}
    \toprule
    \textbf{[Role assignment]} \\
    You are optimizing the policy for a text-to-SQL agent to reduce errors in generating correct SQL queries. \\
    \\
    \textbf{[Reference materials]} \\
    Here are several error cases that have occurred: \\
    \texttt{"''\textcolor{blue}{\{previous wrong cases\}}"''} \\
    \\
    Here are several correct cases that have occurred: \\
    \texttt{"''\textcolor{blue}{\{previous correct cases\}}"''} \\
    \\
    \textbf{[Additional instructions]} \\
    Please analyze all the cases and revise the policy accordingly. Focus on text-to-SQL correctness and how to pass the required tests. \\
    Remove redundant or obsolete points. If certain previous policy points are still valid, keep them. \\
    Output the revised policy, starting with POLICY: on a new line, followed immediately by a bulleted list (each point on a new line, starting with - ). \\
    Limit the revised policy to at most 5 concise and actionable bullet points relevant to common text-to-SQL mistakes. \\
    \\
    Example output format: \\
    \texttt{POLICY:} \\
    \texttt{- [first point]} \\
    \texttt{- [second point]} \\
    \bottomrule
  \end{tabular}
  \vspace{-0.7em}
  \caption{The prompt template for the policy generator in text-to-SQL.}
  \label{fig:policy_text2sql}
\end{figure}

\begin{figure}[t!]
  \centering
  \small
  \renewcommand{\arraystretch}{1.1}
  \begin{tabular}{p{0.93\textwidth}}
    \toprule
    \textbf{[Role assignment]} \\
    You are optimizing a QA agent's policy to reduce errors. \\
    \\
    \textbf{[Reference materials]} \\
    Here are several error cases that have occurred: \\
    \texttt{"''\textcolor{blue}{\{previous wrong cases\}}"''} \\
    \\
    Here are several correct cases that have occurred: \\
    \texttt{"''\textcolor{blue}{\{previous correct cases\}}"''} \\
    \\
    \textbf{[Additional instructions]} \\
    Please analyze all the cases and revise the policy accordingly. Focus on question answering correctness and how to pass the required tests. \\
    Remove redundant or obsolete points. If certain previous policies are still valid, keep them. \\
    Output the revised policy, starting with POLICY: on a new line, followed immediately by a bulleted list (each point on a new line, starting with - ). \\
    Limit the revised policy to at most 5 concise, actionable bullet points relevant to common QA mistakes. \\
    \\
    Example output format: \\
    \texttt{POLICY:} \\
    \texttt{- [first point]} \\
    \texttt{- [second point]} \\
    \bottomrule
  \end{tabular}
  \vspace{-0.7em}
  \caption{The prompt template for the policy generator in HotpotQA.}
  \label{fig:policy_hotpotqa}
\end{figure}

\begin{figure}[t!]
  \centering
  \small
  \renewcommand{\arraystretch}{1.1}
  \begin{tabular}{p{0.93\textwidth}}
    \toprule
    \textbf{[Role assignment]} \\
    You are optimizing the policy for a medical QA agent to reduce diagnostic errors. \\
    \\
    \textbf{[Reference materials]} \\
    Here are several error cases that have occurred: \\
    \texttt{"''\textcolor{blue}{\{previous wrong cases\}}"''} \\
    \\
    Here are several correct cases that have occurred: \\
    \texttt{"''\textcolor{blue}{\{previous correct cases\}}"''} \\
    \\
    \textbf{[Additional instructions]} \\
    Please analyze all the cases and revise the policy accordingly. Focus on diagnosis correctness and how to pass the required tests. \\
    Remove redundant or obsolete points. If certain previous policies are still valid, keep them. \\
    Output the revised policy, starting with POLICY: on a new line, followed immediately by a bulleted list (each point on a new line, starting with - ). \\
    Limit the revised policy to at most 5 concise, actionable bullet points relevant to common diagnostic mistakes. \\
    \\
    Example output format: \\
    \texttt{POLICY:} \\
    \texttt{- [first point]} \\
    \texttt{- [second point]} \\
    \bottomrule
  \end{tabular}
  \vspace{-0.7em}
  \caption{The prompt template for the policy generator in DDXPlus.}
  \label{fig:policy_ddxplus}
\end{figure}

\begin{figure}[t!]
  \centering
  \small
  \renewcommand{\arraystretch}{1.1}
  \begin{tabular}{p{0.93\textwidth}}
    \toprule
    \textbf{[Role assignment]} \\
    You are optimizing the policy for a Python programming agent to reduce errors in solving user requirements. \\
    \\
    \textbf{[Reference materials]} \\
    Here are several error cases that have occurred: \\
    \texttt{"''\textcolor{blue}{\{previous wrong cases\}}"''} \\
    \\
    Here are several correct cases that have occurred: \\
    \texttt{"''\textcolor{blue}{\{previous correct cases\}}"''} \\
    \\
    \textbf{[Additional instructions]} \\
    Please analyze all the cases and revise the policy accordingly. Focus on Python code correctness and how to pass the required tests. \\
    Remove redundant or obsolete points. If certain previous policies are still valid, keep them. \\
    Output the revised policy, starting with POLICY: on a new line, followed immediately by a bulleted list (each point on a new line, starting with - ). \\
    Limit the revised policy to at most 5 concise, actionable bullet points relevant to common Python coding and testing mistakes. \\
    \\
    Example output format: \\
    \texttt{POLICY:} \\
    \texttt{- [first point]} \\
    \texttt{- [second point]} \\
    \bottomrule
  \end{tabular}
  \vspace{-0.7em}
  \caption{The prompt template for the policy generator in DS-1000.}
  \label{fig:policy_ds1000}
\end{figure}

\end{document}